%% file: paper.tex
\documentclass[runningheads]{llncs}
\usepackage{graphicx}
\usepackage{comment}
\usepackage{amsmath,amssymb} %
\usepackage{color}
\usepackage{xcolor}

\usepackage{booktabs}

\usepackage{xspace}

\usepackage{algorithm}
\usepackage{algpseudocode}
\usepackage{multirow}
\usepackage{tabularx}
\usepackage{wrapfig}

\makeatletter
\DeclareRobustCommand\onedot{\futurelet\@let@token\@onedot}
\def\@onedot{\ifx\@let@token.\else.\null\fi\xspace}

\usepackage[pagebackref=true,breaklinks=true,letterpaper=true,colorlinks,bookmarks=false]{hyperref}

\def\eg{\emph{e.g}\onedot} 
\def\ie{\emph{i.e}\onedot}

\def\etal{\emph{et al}\onedot}

\newcommand{\figref}[1]{Fig\onedot~\ref{#1}}

\newcommand{\tabref}[1]{Tab\onedot~\ref{#1}}
\newcommand{\algoref}[1]{Alg\onedot~\ref{#1}}

\usepackage{pifont}
\newcommand{\cmark}{\ding{51}}

\DeclareMathOperator*{\argmin}{arg\,min}

\begin{document}
\pagestyle{headings}
\mainmatter
\def\ECCVSubNumber{942}  %

\title{Naive-Student: Leveraging Semi-Supervised Learning in Video Sequences for Urban Scene Segmentation}

\titlerunning{Semi-Supervised Learning in Video Sequences for Urban Scene Segmentation} 
\authorrunning{L-C Chen \etal} 
\author{Liang-Chieh~Chen\inst{1},
Raphael~Gontijo~Lopes\inst{1}, 
Bowen~Cheng\inst{2},
Maxwell~D.~Collins\inst{1},
Ekin~D.~Cubuk\inst{1},
Barret~Zoph\inst{1}, \\
Hartwig~Adam\inst{1},
Jonathon~Shlens\inst{1}}
\institute{$^1$ Google Research, $^2$ UIUC}

\maketitle

\input{sections/0.abstract.tex}
\input{sections/1.intro.tex}
\input{sections/2.related.tex}
\input{sections/3.method.tex}

\input{sections/4.experiments.tex}
\input{sections/5.conclusion.tex}

\clearpage
\bibliographystyle{splncs04}
\bibliography{refs}

\end{document}

%% file: sections/0.abstract.tex
\begin{abstract}

Supervised learning in large discriminative models is a mainstay for modern computer vision. Such an approach necessitates investing in large-scale human-annotated datasets for achieving state-of-the-art results. In turn, the efficacy of supervised learning may be limited by the size of the human annotated dataset. This limitation is particularly notable for image segmentation tasks, where the expense of human annotation is especially large, yet large amounts of unlabeled data may exist. In this work, we ask if we may leverage semi-supervised learning in unlabeled video sequences and extra images to improve the performance on urban scene segmentation, simultaneously tackling semantic, instance, and panoptic segmentation. The goal of this work is to avoid the construction of sophisticated, learned architectures specific to label propagation (\eg, patch matching and optical flow). Instead, we simply predict pseudo-labels for the unlabeled data and train subsequent models with both human-annotated and pseudo-labeled data. The procedure is iterated for several times. As a result, our Naive-Student model, trained with such simple yet effective iterative semi-supervised learning, attains state-of-the-art results at all three Cityscapes benchmarks, reaching the performance of 67.8\% PQ, 42.6\% AP, and 85.2\% mIOU on the test set. We view this work as a notable step towards building a simple procedure to harness unlabeled video sequences and extra images to surpass state-of-the-art performance on core computer vision tasks.

\keywords{semi-supervised learning, pseudo label, semantic segmentation, instance segmentation, panoptic segmentation}
\end{abstract}

%% file: sections/1.intro.tex
\begin{figure}
    \vspace{-0.32in}
    \centering
    \begin{tabular}{c c c c}
        \includegraphics[width=0.56\textwidth]{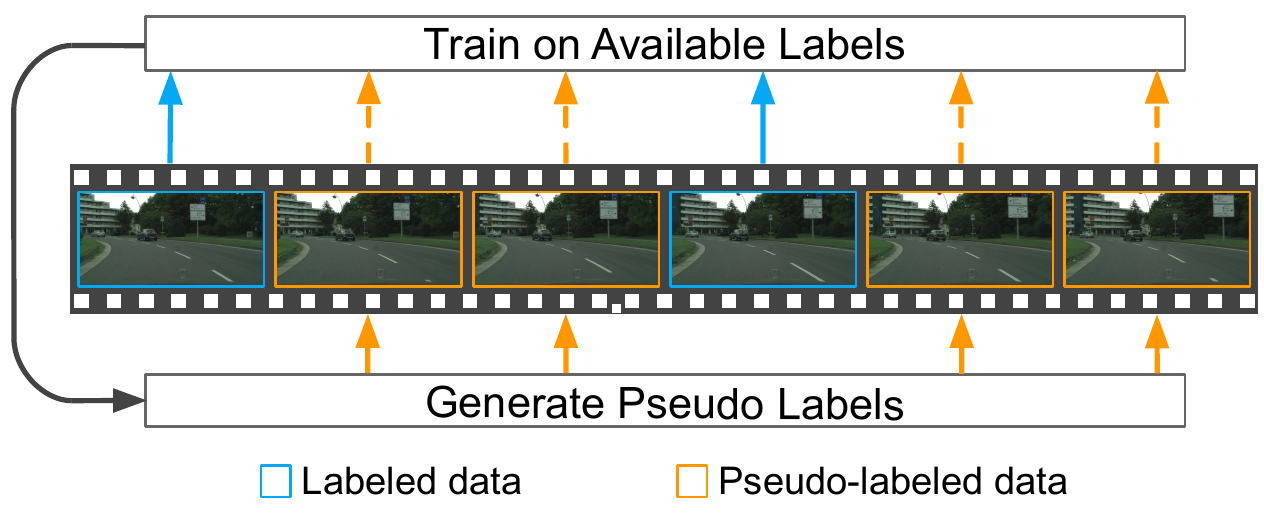}  %
        & \hspace{0.15in} 
        & %
        \includegraphics[width=0.35\textwidth]{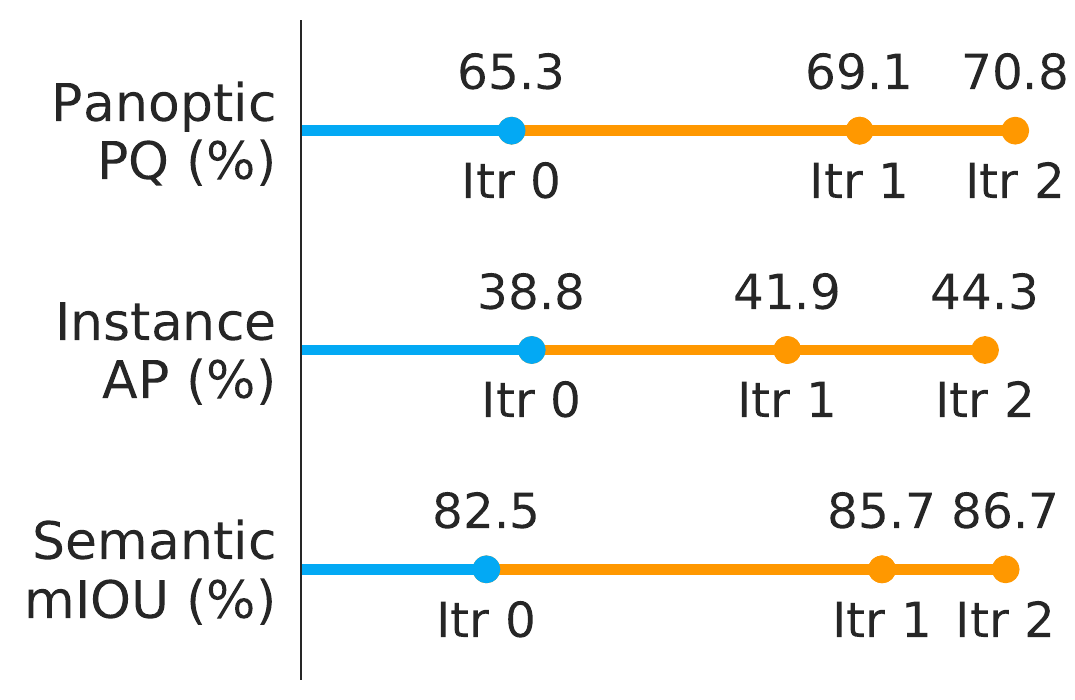}
        & \hspace{0.15in} 
    \end{tabular}%
    \vspace{-0.05in}
    \caption{{\bf Naive-Student: semi-supervised learning in video sequences for scene segmentation}. We iteratively train on human-annotated frames (1 out of 30 in each Cityscapes video sequence), and generate pseudo-labels for the other unlabeled video frames (left). Segmentation performances (val set) improve at each iteration (right).}
    \vspace{-0.5in}
    \label{fig:header}
\end{figure}

\section{Introduction}
\label{sec:intro}
Significant advances in computer vision due to deep learning \cite{krizhevsky2012imagenet,ren2015faster,girshick2014rich,deeplabv12015} have been tempered by the fact that these advances have been accrued through supervised learning on large-scale, human-annotated datasets \cite{lin2014microsoft,ILSVRC15}.
The paradigm of supervised learning requires the expenditure of a large amount of resources to manually label static images -- whether through the development of specialized annotation tools \cite{bell2013opensurfaces,russell2008labelme,castrejon2017annotating}, or the amount of human hours for the annotation itself \cite{lin2014microsoft,ILSVRC15}.
Such an approach does not scale effectively to comprehensively label real-time video frames (but see \cite{real2017youtube,abu2016youtube,caba2015activitynet}).
More importantly, supervised training is rather sample-inefficient as many examples are required for good generalization \cite{henaff2019data,kornblith2019better}.
Ideally, one would expect and hope that a training method may be able to learn in a more self-supervised manner particularly on video -- much as presumed to occur in human visual learning \cite{lake2017building,wu2015galileo}.

The limitations of supervised learning is most pronounced in the task of image segmentation \cite{forsyth2002computer}.
Human annotation of static images for segmentation is particularly expensive, requiring, for instance, 90 minutes per image \cite{cordts2016cityscapes} or 22 worker hours per 1,000 mask segmentations \cite{lin2014microsoft}.
In the case of self-driving cars, the annotation of video is a critical supervised learning problem \cite{geiger2013vision,sun2019scalability}, and in turn has fostered an industry of specialized companies for data annotation.

In contrast, recent findings on the benefits of pre-training on ImageNet for segmentation \cite{li2017fully} indicate that current segmentation approaches may benefit from large-scale image classification datasets. This direction has been further pursued by \cite{sun2017revisiting,chen2017deeplabv3} on an extremely large image classification dataset \cite{hinton2015distilling}. Additionally, many segmentation methods \cite{zhao2017pyramid,chen2018searching} apply transfer learning by pre-training on augmented segmentation datasets \cite{BharathICCV2011,lin2014microsoft,neuhold2017mapillary} and then fine-tuning on the target datasets \cite{everingham2010pascal,cordts2016cityscapes}. Likewise, other works attempt to exploit label propagation in video to improve  segmentation. However, these methods require building specialized modules to propagate labels across video frames \cite{mustikovela2016can,gadde2017semantic,nilsson2018semantic,zhu2019improving}.

In this work, we leverage both unlabeled video frames and extra unlabeled images  to improve the urban scene segmentation evaluated in terms of semantic segmentation, instance segmentation, and panoptic segmentation. Importantly, we do not require any specialized methods for propagating label information across video frames, such as optical flow \cite{mustikovela2016can,gadde2017semantic,nilsson2018semantic}, patch matching \cite{badrinarayanan2010label,budvytis2017large}, or learned motion vector \cite{zhu2019improving}. Instead, we propose to employ a simple iterative semi-supervised learning procedure. At each iteration, the model from the previous iteration generates pseudo-labels for unlabeled video frames (Figure \ref{fig:header}).
Specifically, a pseudo-label is generated through a distillation across multiple augmentations applied to each unlabeled video frame.
Subsequent iterations of the training procedure train on the original labeled data as well as the newly pseudo-labeled data. Our model, trained with such a simple yet effective method, simultaneously sets new state-of-the-art results on the Cityscapes urban scene segmentation \cite{cordts2016cityscapes}, achieving 67.8\% PQ, 42.6\% AP, and 85.2\% mIOU on test set. We hope that such an iterative semi-supervised learning may provide more label-efficient methods for developing a machine learning solution to segmentation.

%% file: sections/2.related.tex
\section{Related Works}
\label{sec:related}

Our method is related to both {\it self-training}~\cite{scudder1965probability,yarowsky1995unsupervised,riloff2003learning,doersch2015unsupervised,predicting_rotation,zhai2019s4l}, where the predictions of a model on unlabeled data is used to train the model, and {\it semi-supervised learning}~\cite{rosenberg2005semi,li2010optimol,papandreou2015weakly,radosavovic2018data,billion_large_scale}, where additionally extra human-annotated data is available to guide the training with unlabeled data. In particular, our model is trained with some human-annotated images and abundant pseudo-labeled~\cite{lee2013pseudo,shi2018transductive,iscen2019label,arazo2019pseudo} video sequences.

Semi-supervised learning has been widely applied to several computer vision tasks, including semantic segmentation~\cite{papandreou2015weakly,dai2015boxsup,pathak2015constrained,hong2015decoupled,wei2016stc,khoreva2017simple,wei2018revisiting,souly2017semi,zhu2020improving}, object detection~\cite{rosenberg2005semi,tang2016large,radosavovic2018data}, instance segmentation~\cite{khoreva2017simple,pinheiro2015learning}, panoptic segmentation~\cite{li2018weakly}, human pose estimation~\cite{papandreou2018personlab}, person re-identification~\cite{zheng2017unlabeled}, multi-object tracking and segmentation \cite{voigtlaender2019mots,porzi2020learning}, and so on. A comprehensive literature survey is beyond the scope of this work, and thus we focus on comparing our proposed method with the most related ones.

Our proposed iterative semi-supervised learning is similar to the work by Papandreou~\etal~\cite{papandreou2015weakly}, STC~\cite{wei2016stc}, Simple-Does-It~\cite{khoreva2017simple}, the work by Li~\etal~\cite{li2018weakly}, and Noisy-Student~\cite{xie2019self}. In particular, our iterative semi-supervised learning is similar to the Expectation-Maximization method by Papandreou~\etal~\cite{papandreou2015weakly} which alternates between estimating the latent pixel labels (\ie, pseudo labels) and optimizing the network parameters with bounding box or image-level annotations. Similarly, Li~\etal~\cite{li2018weakly} generate pseudo labels for panoptic segmentation by exploiting both fully-annotated and weakly-annotated images, where bounding boxes for `thing' classes and image-level tags for `stuff' classes are provided. However, unlike those two works, we do not exploit any weakly-annotated data. Additionally, we do not sort the images by the annotation difficulty and do not exploit any other assistance, such as saliency maps, as in STC~\cite{wei2016stc}. Simple-Does-It~\cite{khoreva2017simple} adopts a complicated de-noising procedure to clean the pseudo labels, while we simply use the outputs from a neural network. Finally, following Noisy-Student~\cite{xie2019self}, we employ a stronger Student network in the subsequent iterations, but we do not employ any noisy data augmentation (\ie, RandAugment~\cite{cubuk2019randaugment}).

When generating pseudo labels, we employ a simple test-time augmentation, \ie, multi-scale inputs and left-right flips, a common strategy used by segmentation models \cite{chen2017deeplabv3,zhao2017pyramid}, which bears a similarity to Data-Distillation \cite{radosavovic2018data}. However, our framework is deployed in an iterative manner, and we exploit unlabeled video sequences for scene segmentation, simultaneously tackling semantic, instance, and panoptic segmentation. Additionally, we do not set a threshold as \cite{radosavovic2018data} to remove false positives, avoiding tuning of another hyper-parameter.

Video sequences have also been exploited in semi-supervised learning for semantic segmentation. Human-annotated ground-truth labels of certain frames in a video sequence could be propagated to other unlabeled frames via patch matching~\cite{badrinarayanan2010label,budvytis2017large} or optical flow~\cite{mustikovela2016can,gadde2017semantic,zhu2017deep,luc2017predicting,nilsson2018semantic}. Recently, Zhu \etal~\cite{zhu2019improving} generate pseudo-labeled video sequences by jointly propagating the image-label pair with learned motion vectors, and demonstrate promising results. Similarly, our method also exploits unlabeled video sequences. However, our method is much simpler since we do not employ any label-propagation modules (\eg, patch matching \cite{badrinarayanan2010label,budvytis2017large}, optical flow \cite{mustikovela2016can,gadde2017semantic,nilsson2018semantic}, or motion vectors \cite{zhu2019improving}) but instead directly generate the pseudo labels for each video frame.

%% file: sections/3.method.tex
\section{Methods}
\label{sec:methods}

\begin{figure}[!t]
    \centering
    \includegraphics[width=1\textwidth]{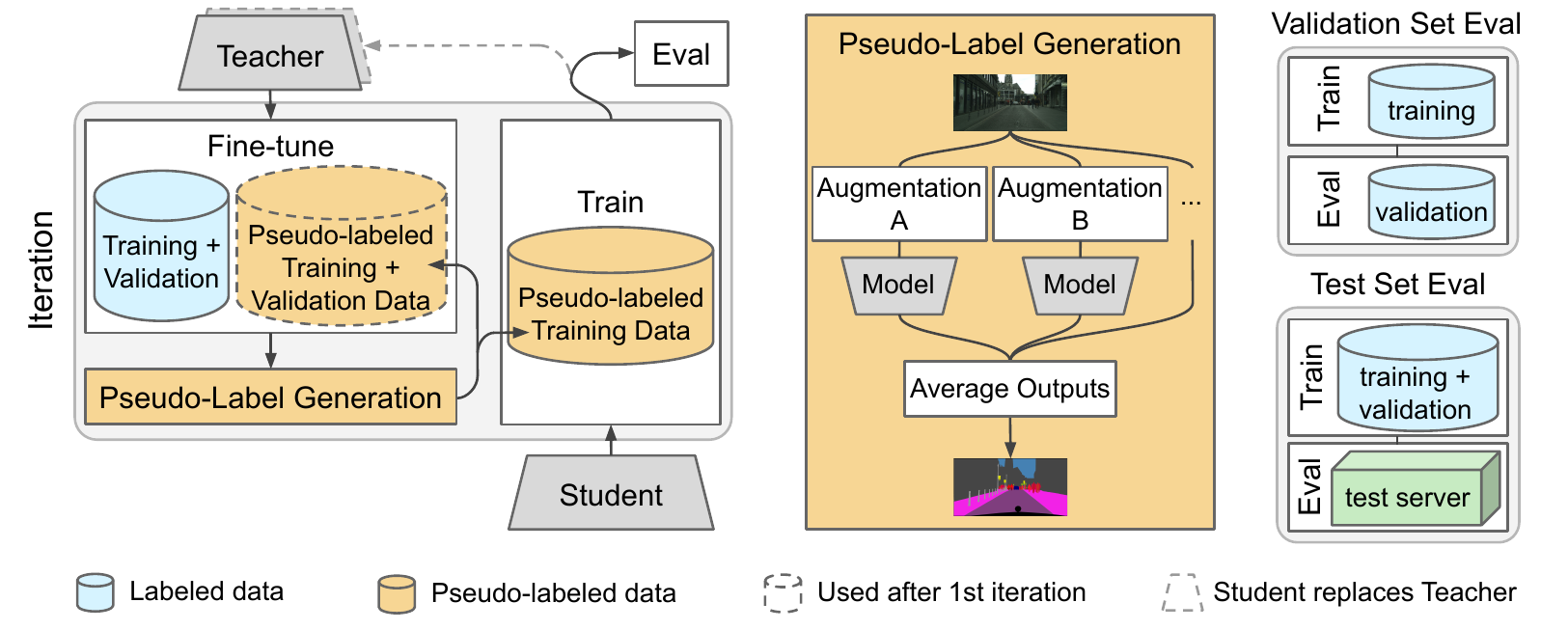}
    \caption{{\bf Overview of our proposed iterative semi-supervised learning for scene segmentation.} The Teacher network is trained with all available human-annotated images (and extra pseudo-labeled images after 1st iteration), and then generates pseudo-labels for all the unlabeled images with a simple test-time augmentation (\ie, multi-scale inputs and left-right flips). The Student network is subsequently trained with the pseudo-labeled data, and optionally replaces the Teacher network in following iterations. Before evaluating the validation or test set performance, the Student network is fine-tuned on the human-annotated images. Note that the validation set is only exploited by the Teacher in order to generate high-quality pseudo-labels, and the Student has no access to it. Additionally, the final test set results are evaluated on a fair test server where the annotations are held-out.
    }
    \label{fig:methods}
\end{figure}

\algoref{alg:semi} gives an overview of our proposed iterative semi-supervised learning for scene segmentation. Suppose two sets of images are given, where one contains human annotations and the other does not. The human-annotated images are exploited to train a Teacher network using the loss function for scene segmentation. Pseudo-labels for those un-annotated images are then generated by the Teacher network with a test-time augmentation function. A Student network is subsequently trained with the pseudo-labeled images using the same loss function for scene segmentation. The Student network is then fine-tuned on human-labeled images before evaluating on the validation set or test set. Finally, one could optionally replace the Teacher network with the Student network and iterate the procedure again. Our method, dubbed {\it Naive-Student}, is motivated by Noisy-Student~\cite{xie2019self} where we adopt a stronger Student network in the following iterations, but we do not inject noise (\ie, RandAugment~\cite{cubuk2019randaugment}) to the Student. Our algorithm is illustrated in \figref{fig:methods}. We elaborate on the details below. 

\begin{algorithm}[!t]
\centering
\caption{Iterative semi-supervised learning for urban scene segmentation.}
\label{alg:semi}
\resizebox{1.0\textwidth}{!}{%
\begin{minipage}{1.4\textwidth}
\begin{algorithmic}
\State \textbf{Labeled data}: $n$ pairs of image $x_i$ and corresponding human annotation $y_i$
\State \textbf{Unlabeled data}: $m$ images collected from multiple video sequences or extra images with no human annotations $\{\widetilde{x}_1, \widetilde{x}_2, ..., \widetilde{x}_m\}$.
\State \textbf{Step 1}: Train a Teacher network $\theta_{t}$ (with prediction function $f$) on the manually labeled images by minimizing the total loss $\mathcal{L}$ for scene segmentation.
$$\theta_{t}^{*} = \argmin_{\theta_t}\frac{1}{n}\sum_{i=1}^{n}\mathcal{L}(y_i, f(x_i, \theta_{t}))$$
where $\mathcal{L} = \lambda_\text{sem}\mathcal{L}_\text{sem} + \lambda_\text{heatmap}\mathcal{L}_\text{heatmap} + \lambda_\text{offset}\mathcal{L}_\text{offset}$ in our framework.
\State \textbf{Step 2}: Generate pseudo-labels $\widetilde{y}_i$ for unlabeled images with test-time augmentations (\ie, multi-scale inputs and left-right flips).

$$\widetilde{y}_i = f(\text{\it Aug}\,(\,\widetilde{x}_i), \theta_{t}^{*}), \forall i = 1, ..., m$$
where $\text{\it Aug}\,(\cdot)$ is test-time augmentation.

\State \textbf{Step 3}: Train an \textit{equal or larger} Student network $\theta_{s}$ on pseudo-labeled images ($\widetilde{x}_i$, $\widetilde{y}_i$) with the same objective.
$$\theta_{s}^{*} = \argmin_{\theta_s}\frac{1}{m}\sum_{i=1}^{m}\mathcal{L}(\widetilde{y}_i, f(\widetilde{x}_i, \theta_{s}))$$
\State \textbf{Step 4}: Fine-tune the Student network $\theta_{s}^{*}$ from step 3 on the manually labeled image annotations ($x_i, y_i$) using the same objective.
$$\theta_{s}^{**} = \argmin_{\theta_s^{*}}\frac{1}{n}\sum_{i=1}^{n}\mathcal{L}({y}_i, f({x}_i, \theta_{s}^{*}))$$
\State \textbf{Step 5}: Return to step 2 but employ the Student network $\theta_{s}^{**}$ as a Teacher until reaching desired number of iterations.
\end{algorithmic}
\end{minipage}%
}%
\end{algorithm}

{\bf The Loss for Scene Segmentation:} Our core building block is the state-of-the-art bottom-up panoptic segmentation model, Panoptic-DeepLab~\cite{cheng2019panopticworkshop}, which improves the semantic segmentation model DeepLabv3+~\cite{deeplabv3plus2018} by incorporating another class-agnostic instance segmentation prediction. Its instance segmentation prediction involves a simple instance center prediction as well as the offset regression from each pixel to its corresponding center. As a result, the total loss function $\mathcal{L}$ for scene segmentation boils down to three loss functions: softmax cross entropy loss $\mathcal{L}_{sem}$ for semantic segmentation, mean squared error loss $\mathcal{L}_{heatmap}$ for instance center prediction, and $L_1$ loss $\mathcal{L}_{offset}$ for offset regression. In our algorithm, the Teacher and the Student networks are trained with the same total loss function $\mathcal{L}$.

{\bf Pseudo-Label Generation:} After training the Teacher network on all human-annotated images (and all pseudo-labeled images after iteration 1), we generate (or update) the pseudo labels for all un-annotated images with a test-time augmentation function $\text{\it Aug}\,(\cdot)$. We simply use the common test-time augmentations, \ie, multi-scale inputs and left-right flips. We only generate hard pseudo labels (\ie, a one-hot distribution) in order to save disk space when processing large resolution images (\eg, Cityscapes image size is $1024\times2048$).

{\bf Ego-Car Region in Pseudo Labels:} Cityscapes images are collected (or recorded) with a driving vehicle. A part of the vehicle, called "ego-car" region, is thus visible in all frames of a video sequence. This region is ignored during evaluating the model performance. However, we find that assigning a random pseudo label value to those regions will confuse models during training. To handle this problem, we adopt a simple solution by exploiting the prior that Cityscapes images are all well-calibrated and the ego-car regions are in the same locations for images collected from the same sequence. Since we have access to the only one human-annotated image from a 30-frame sequence, we propagate this ego-car region information to the other 29 frames in the same sequence and assign them with void label (\ie, no loss back-propagation for those regions).

{\bf A Better Network Backbone for Scene Segmentation:} The efficient backbone Xception-71 (X-71)~\cite{chollet2017xception,dai2017coco,deeplabv3plus2018} is adopted in the Teacher network at the first iteration in our iterative semi-supervised learning algorithm. In the next iteration, a stronger backbone should be used to generate pseudo labels with a better quality. In this work, we modify the powerful Wide ResNet-38 (WR-38) \cite{wu2019wider,wrn2016wide} for scene segmentation. In particular, we remove the last residual block B7 in WR-38~\cite{wu2019wider} and repeat the residual block B6 two more times, resulting in our proposed WR-41. Additionally, we adopt drop path~\cite{huang2016deep,liu2019auto} (with a constant survival probability 0.8) and multi-grid scheme \cite{chen2017deeplabv3,wang2017understanding} in the last three residual blocks (with unit rate $\{1, 2, 4\}$, same as \cite{chen2017deeplabv3}). As a result, the proposed WR-41 attains better performance than X-71 in the fully supervised setting. %

%% file: sections/4.experiments.tex
\section{Experiments}
\label{sec:experiments}

We conduct experiments on the popular Cityscapes dataset \cite{cordts2016cityscapes}, which consists of a large and diverse set of street-view video sequences recorded from 50 cities primarily in Germany. From the video sequences, 5000 images are provided with high-quality pixel-wise annotations in which 2975, 500, and 1525 images are used for training, validation, and test, respectively. Each image is selected from the 20th frame of a 30-frame video snippet. Additionally, another 20000 images are accompanied with coarse annotations. We define each dataset split below.

{\bf train-fine:} Training set (2,975 images) with fine pixel-wise annotations.

{\bf val-fine:} Validation set (500 images) with fine pixel-wise annotations.

{\bf test-fine:} Test set (1,525 images) where the fine pixel-wise annotations are held-out, and the evaluation is performed on a fair test server.

{\bf train-extra:} Extra 20,000 images with coarse annotations. Our proposed method is not limited to video sequences, and thus we also generate pseudo-labels for this set, instead of using the provided coarse annotations.

{\bf train-sequence:} The video sequences where the train-fine set is selected from. This set contains $2975\times30=89,250$ frames.

{\bf val-sequence:} The video sequences where the val-fine set is selected from. This set contains $500\times30=15,000$ frames.

Furthermore, one could merge training and validation splits (\eg, trainval-fine is merged from train-fine and val-fine, and similarly for trainval-sequence).

{\bf Experimental Setup:} We report mean intersection-over-union (mIOU), average precision (AP), and panoptic quality (PQ) to evaluate the semantic, instance, and panoptic segmentation results, respectively.

The state-of-art bottom-up panoptic segmentation model, Panoptic-DeepLab \cite{cheng2019panoptic}, is included in our proposed iterative semi-supervised learning pipeline. Panoptic-DeepLab is a simple framework and simultaneously produces semantic, instance, and panoptic segmentation results without the need to fine-tune on each task. We adopt the same training protocol as \cite{cheng2019panoptic} when using Panoptic-DeepLab. For example, our models are trained using TensorFlow \cite{tensorflow-osdi2016} on 32 TPUs. We use the `poly' learning rate policy~\cite{liu2015parsenet} with an initial learning rate of $0.001$ for Xception-71 (X-71) backbone \cite{chollet2017xception,dai2017coco} and $0.0001$ for our proposed Wide ResNet-41 (WR-41) \cite{he2016deep,wrn2016wide,wu2019wider}, respectively. During training, the batch normalization~\cite{ioffe2015batch} is fine-tuned, random scale data augmentation and Adam \cite{kingma2015adam} optimizer without weight decay are adopted. On Cityscapes, we employ training crop size equal to $1025\times2049$ with batch size 32, and 180K training iterations. Similar to other works on panoptic segmentation \cite{kirillov2018panoptic,li2018learning,kirillov2019panoptic,xiong2019upsnet,porzi2019seamless,yang2019deeperlab,wang2020axial}, we re-assign to void label all `stuff` segments whose areas are smaller than a threshold of 4096. Additionally, we employ multi-scale inference (scales equal to $\{0.5, 0.75, 1, 1.25, 1.5, 1.75, 2\}$ for Cityscapes) and left-right flipped inputs, to further improve the performance for test server evaluation.

\subsection{Urban Scene Segmentation Results}

In this subsection, we summarize our main results on the Cityscapes dataset.

{\bf Cityscapes val-fine set:} In our iterative semi-supervised learning framework, at each iteration, all data splits, including Mapillary Vistas \cite{neuhold2017mapillary} and Cityscapes trainval-fine, (also trainval-sequence and train-extra after 1st iteration), are exploited for the Teacher networks in order to generate better pseudo-labels, while the Student networks are always initialized from the Mapillary Vistas pretrained checkpoint (unless it is specified that it is initialized from previous iterations). In \tabref{tab:cityscapes_val}, we report the validation set results. At iteration 0, we employ the state-of-art Panoptic-DeepLab with Xception-71 (validation set results from \cite{cheng2019panoptic} are shown in the table for comparison) as the Teacher network to generate pseudo-labels for train-sequence and train-extra splits which are subsequently used to train our Student network using the proposed Wide ResNet-41 as backbone. As a result, at iteration 1, we improve over the Panoptic-DeepLab (X-71) baseline by a margin of 3.8\% PQ, 3.1\% AP, and 3.2\% mIOU. The Student network is then selected as the new Teacher network after fine-tuning on all the available data splits (\ie, trainval-sequence, train-extra, and trainval-fine). At iteration 2, by training with the better quality pseudo-labels, we observe an additional improvement of 1.3\% PQ, 1.5\% AP, and 0.8\% mIOU for the new Student network. Additionally, one could further slightly improve the performance by initializing the Student network from iteration 1, as shown in the last row.

{\bf Cityscapes test-fine set:} In \tabref{tab:cityscapes_test}, we report our Cityscapes test set results. As shown in the table, our single model simultaneously ranks 1st at all three Cityscapes benchmarks. In particular, for the panoptic segmentation benchmark, our model outperforms Panoptic-DeepLab (X-71) \cite{cheng2019panoptic} by 2.3\% PQ, Li~\etal~\cite{li2020unifying} by 4.5\% PQ, and Seamless~\cite{porzi2019seamless} by 5.2\% PQ. For the instance segmentation benchmark, our model outperforms PolyTransform~\cite{liang2019polytransform} by 2.5\% AP, Panoptic-DeepLab (X-71) \cite{cheng2019panoptic} by 3.6\% AP, and PANet \cite{liu2018path} by 6.2\% AP. Finally, for the competitive semantic segmentation benchmark, our model outperforms Panoptic-DeepLab (X-71) \cite{cheng2019panoptic} by 1.0\% mIOU, OCR~\cite{yuan2019object} by 1.5\% mIOU, and Zhu~\etal~\cite{zhu2019improving} by 1.7\% mIOU.

\begin{table*}[!t]
  \centering
  \caption{{\bf Iterative semi-supervised training systematically improves urban scene segmentation results.} 
  Results presented on Cityscapes validation data. The Students are pretrained on ImageNet \cite{ILSVRC15} and Mapillary Vistas  \cite{neuhold2017mapillary}. Baseline Xception-71 (X-71) at iteration 0 is obtained from \cite{cheng2019panoptic}, while Wide-ResNet-41 (WR-41) is {\it modified} from \cite{wu2019wider}. All Teacher networks have been trained on ImageNet, Mapillary Vistas, and Cityscapes. Labeled data is from Cityscapes train-fine set. Pseudo-Labeled data is from Cityscapes train-sequence and train-extra sets.
  $^\dagger$ indicates that model was initialized from the checkpoint in the previous iteration. The text color indicates how the Student is selected as the new Teacher (\eg, the Student X-71 at iteration 0 becomes the Teacher at iteration 1 after fine-tuning).
  }
  
  \scalebox{0.90}{ %
  \begin{tabular}{c | c  c  | c  c | c c c }
    \toprule[0.2em]
    \multicolumn{1}{c}{} & \multicolumn{2}{c}{Architecture} & \multicolumn{2}{c}{Training Set} &  \multicolumn{3}{c}{Validation Set}\\
    Itr & Student & Teacher & Labeled & Pseudo-Labeled & PQ (\%) & AP (\%) & mIOU (\%) \\
    \toprule[0.2em]
    0 & {\textcolor{blue}{X-71}} & - & \cmark & & 65.3 & 38.8 & 82.5 \\
    1 & {\textcolor{orange}{WR-41}} & {\textcolor{blue}{X-71}} &  & \cmark & 69.1 & 41.9 & 85.7  \\
    2 & WR-41 & {\textcolor{orange}{WR-41}} & & \cmark &  70.4 & 43.4 & 86.5  \\
    2 & WR-41$^\dagger$ & {\textcolor{orange}{WR-41}} & & \cmark & 70.8 & 44.3 & 86.7  \\
     \bottomrule[0.1em]
   \end{tabular}
  }
  \label{tab:cityscapes_val}
\end{table*}

\begin{table*}[!t]
  \centering
  \caption{{\bf Iterative semi-supervised learning achieves state-of-the-art urban scene segmentation results.} Results presented on Cityscapes {\it test-fine} set. {\bf C:} Cityscapes coarse annotation. {\bf V:} Cityscapes video. {\bf MV:} Mapillary Vistas. Note that we do not exploit the train-extra coarse annotations, but instead we generate pseudo-labels for them. 
  }
  \scalebox{0.7}{
  \begin{tabular}{c |  c  | c | c c c }
    \toprule[0.2em]
    Model & Extra Data & Training Method & PQ (\%) & AP (\%) & mIOU (\%) \\
    \toprule[0.2em]
    Naive-Student (ours) & C, V, MV & iterative semi-supervised & 67.8 & 42.6 & 85.2  \\
    \hline \hline
    Seamless~\cite{porzi2019seamless}  & MV & supervised & 62.6 & - & - \\
    Li \etal~\cite{li2020unifying} & COCO & supervised & 63.3 & - & - \\
    Panoptic-DeepLab (X-71)~\cite{cheng2019panoptic}& MV & supervised & 65.5 & 39.0 & 84.2 \\
    PANet~\cite{liu2018path}          & COCO & supervised & - & 36.4 & - \\
    PolyTransform~\cite{liang2019polytransform} & COCO & supervised & - & 40.1 & - \\
    Zhu \etal~\cite{zhu2019improving}  & C, V, MV & semi-supervised & - & - & 83.5 \\
    OCR~\cite{yuan2019object} & C, MV & supervised & - & - & 83.7 \\
    \bottomrule[0.1em]
  \end{tabular}
  }
  \label{tab:cityscapes_test}
\end{table*}

\begin{figure*}[!t]
    \centering
    \scalebox{0.9}{
    \begin{tabular}{c c c}
        \includegraphics[clip,trim={0cm 1cm 0cm 1cm},width=0.33\textwidth]{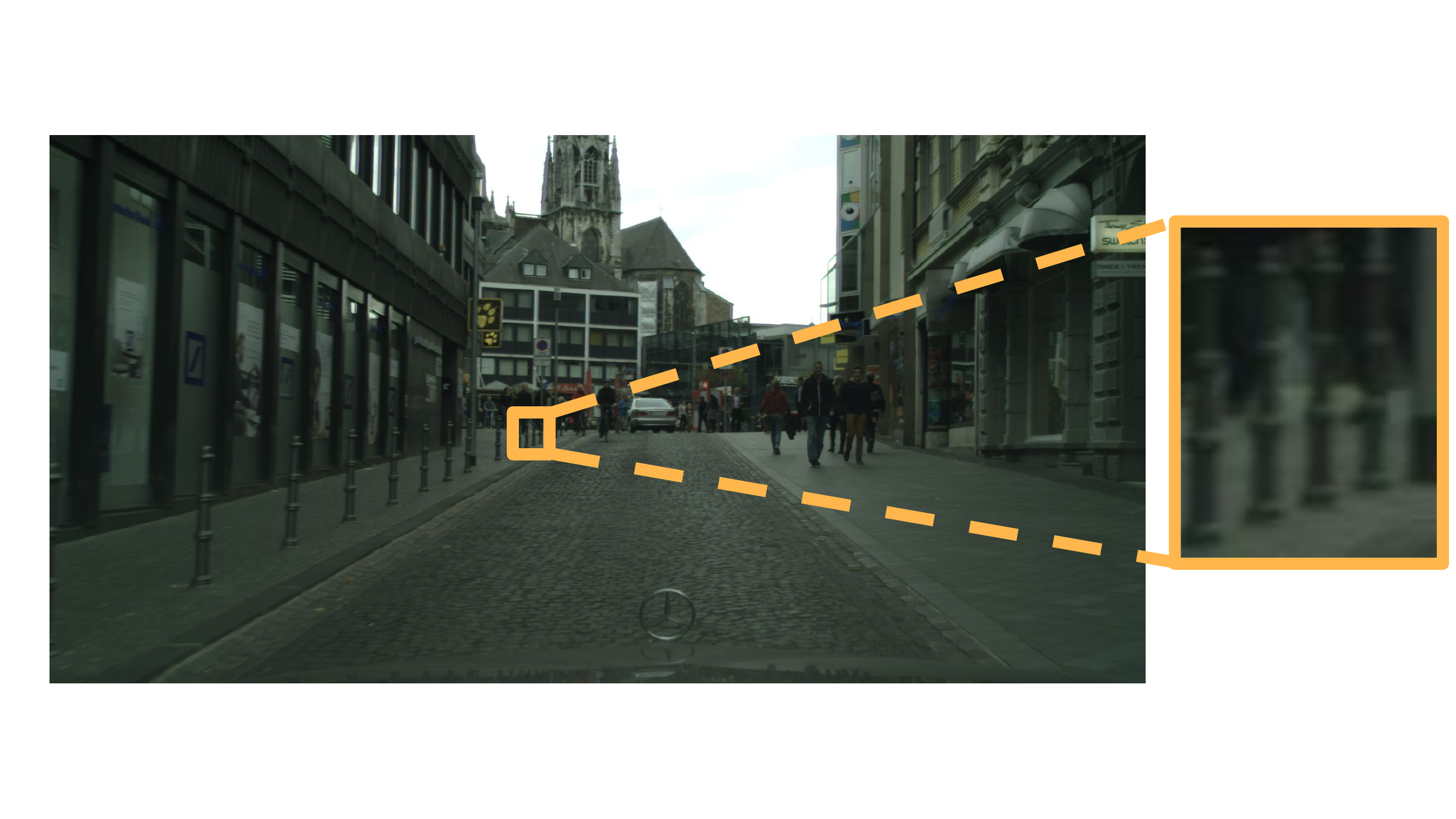} &
        \includegraphics[clip,trim={0cm 1cm 0cm 1cm},width=0.33\textwidth]{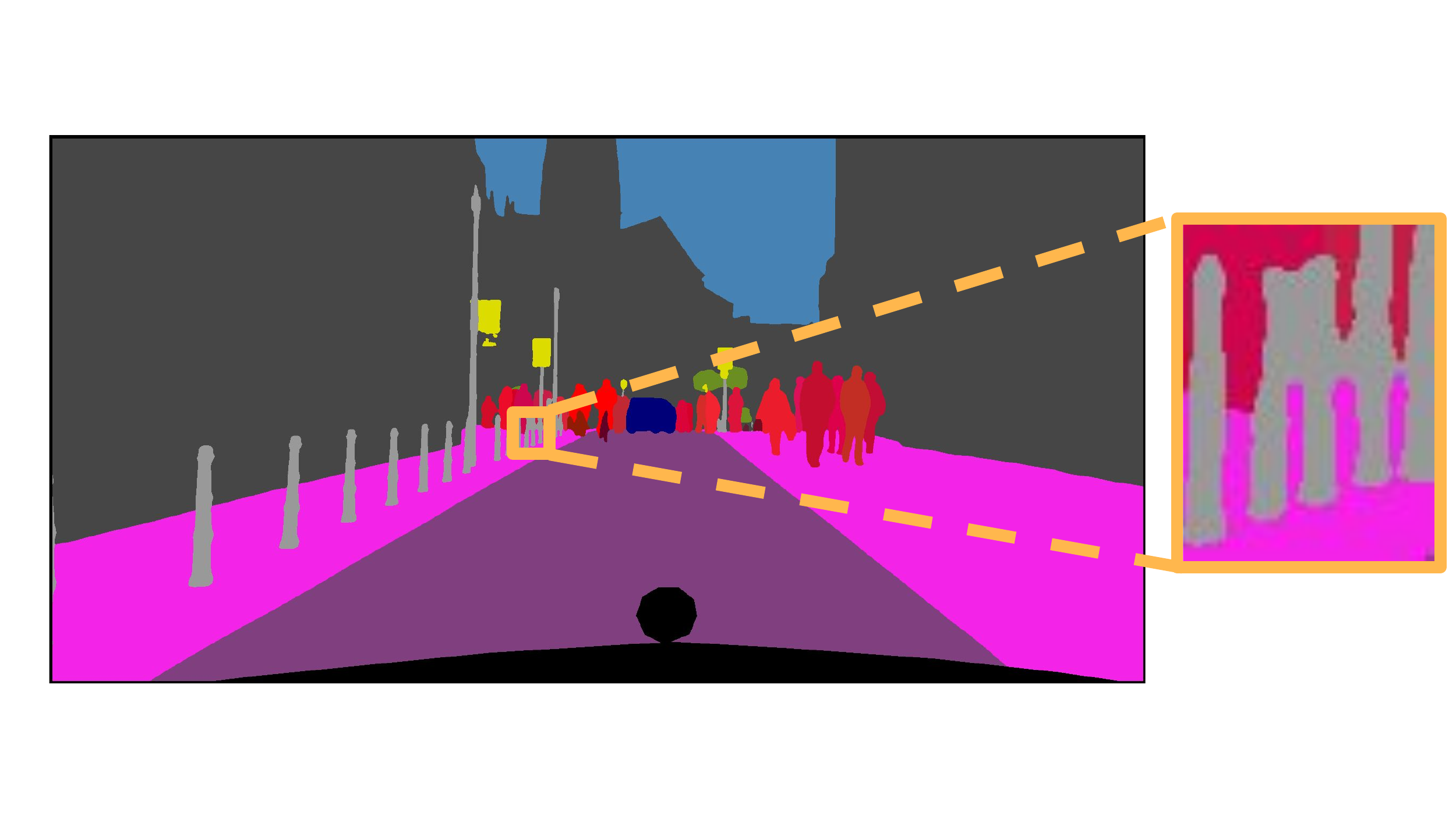} &
        \includegraphics[clip,trim={0cm 1cm 0cm 1cm},width=0.33\textwidth]{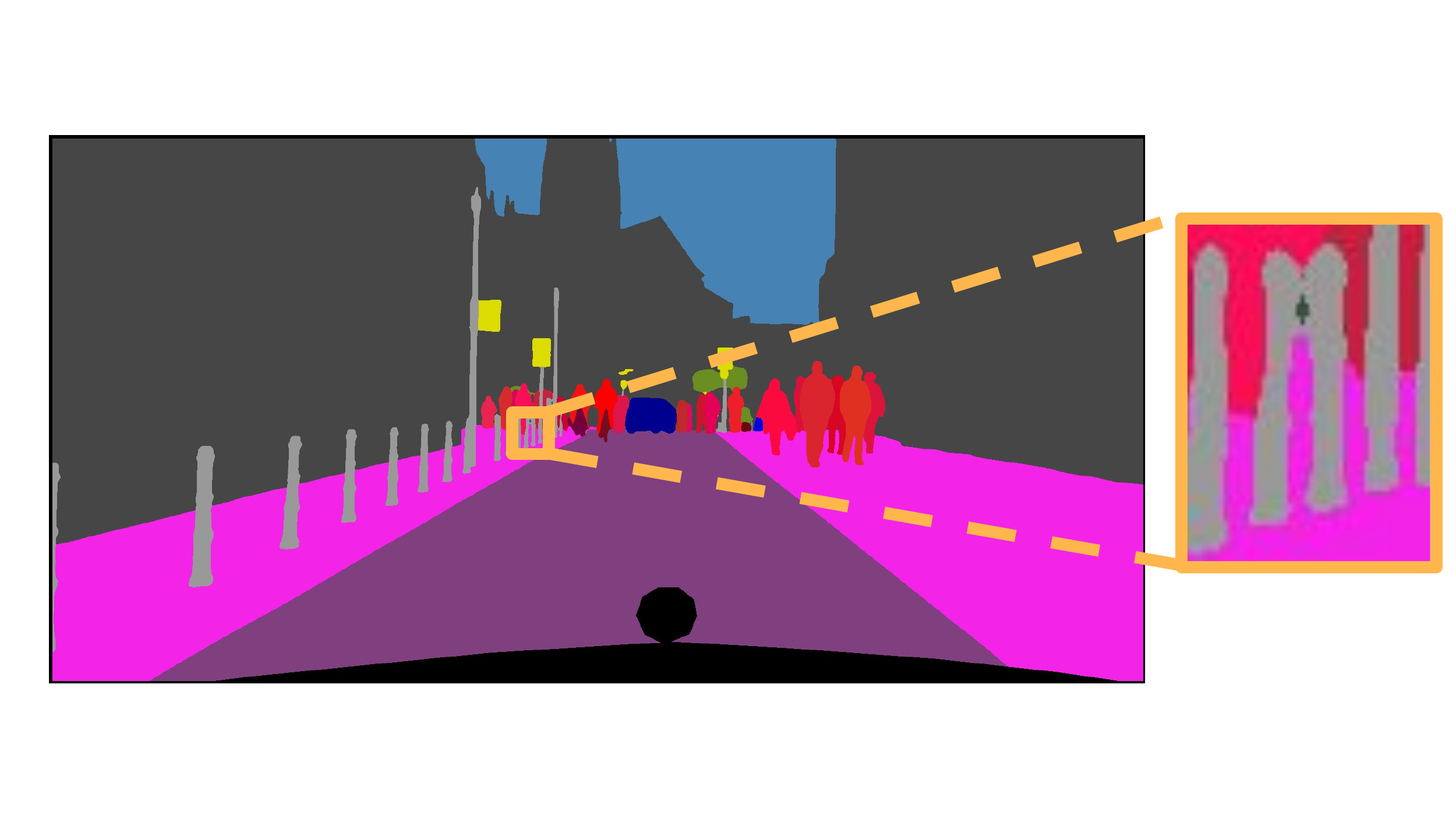} \\
        (a) Image & (b) Pseudo-Label (Itr1) & (c) Pseudo-Label (Itr2) \\
        \includegraphics[clip,trim={0cm 1cm 0cm 1cm},width=0.33\textwidth]{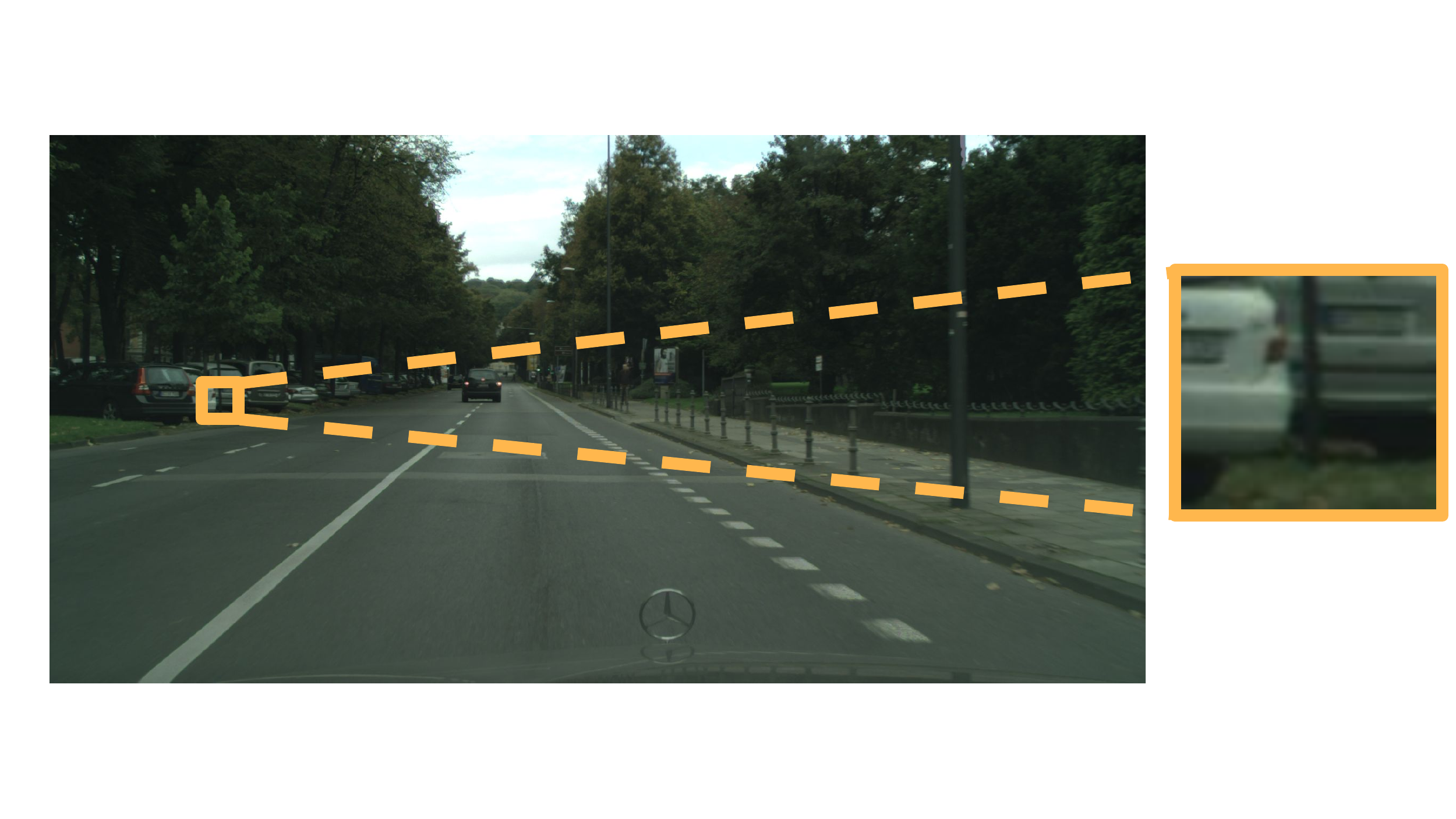} &
        \includegraphics[clip,trim={0cm 1cm 0cm 1cm},width=0.33\textwidth]{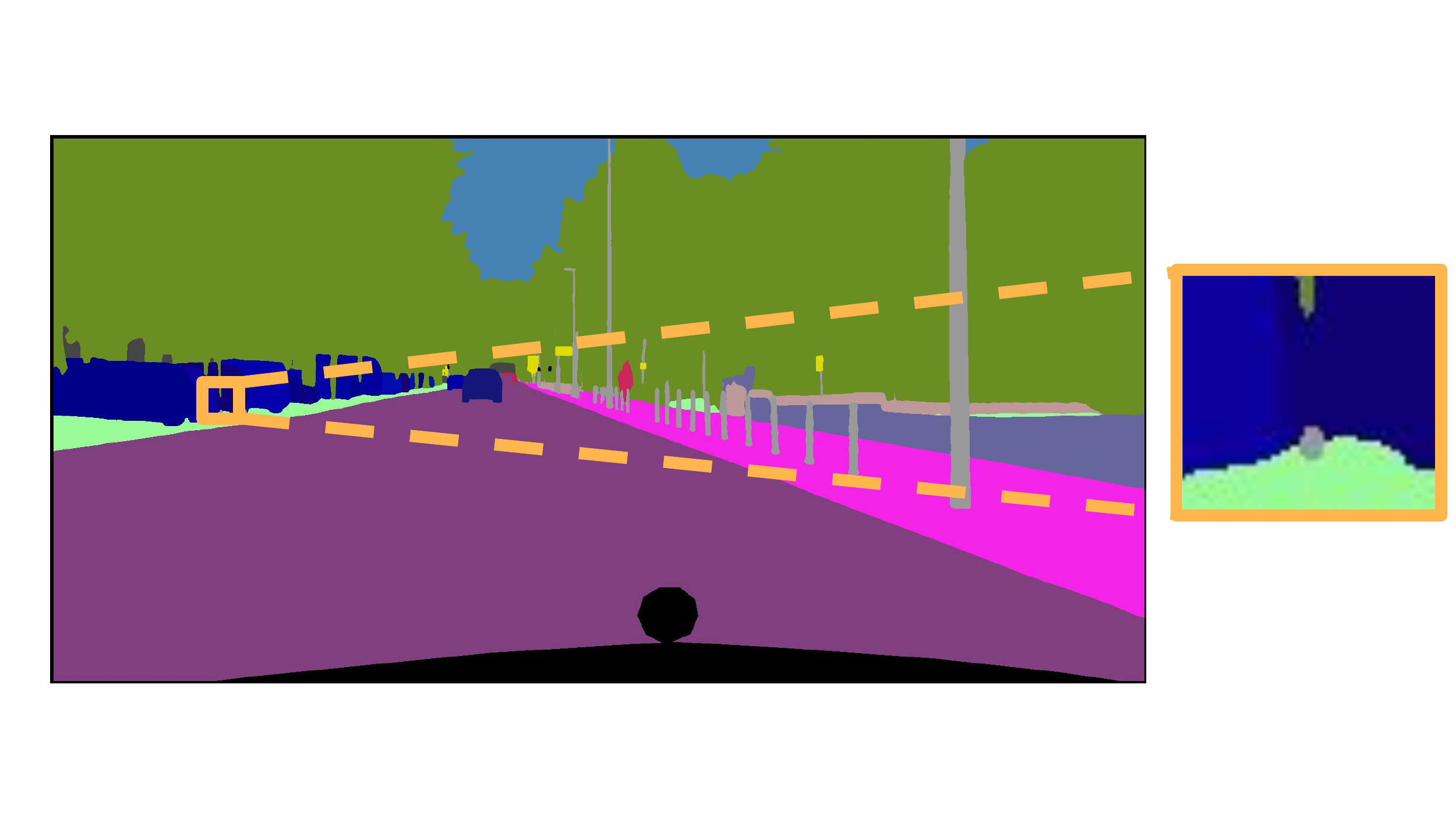} &
        \includegraphics[clip,trim={0cm 1cm 0cm 1cm},width=0.33\textwidth]{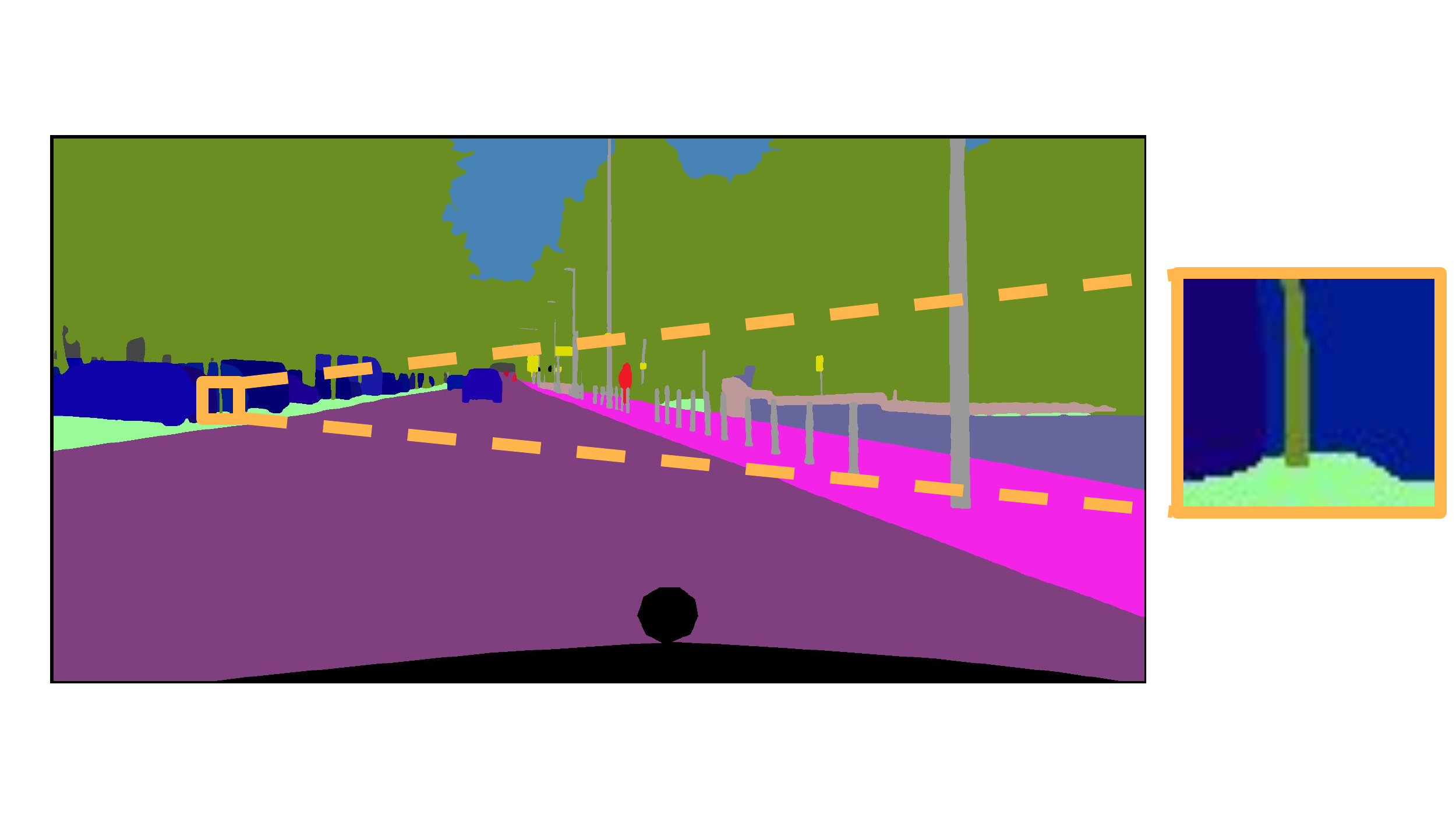} \\
        (a) Image & (b) Pseudo-Label (Itr1) & (c) Pseudo-Label (Itr2) \\
    \end{tabular}
    }
    \caption{{\bf Generated pseudo-labels by Teacher improves qualitatively with more iterations.} We only observe subtle difference between pseudo-labeled video frames at iteration 1 and iteration 2. The results from iteration 2 capture better thin objects, as zoomed-in in the yellow regions. }
    \label{fig:sample_video}
\end{figure*}

{\bf Visualization of generated pseudo labels:} %
We observe visually subtle differences between iteration 1 and iteration 2, since both Teachers yield high-quality results. To further look into the minor differences, we zoom-in some generated pseudo-labels in \figref{fig:sample_video}. As shown in the figure, the Teacher at iteration 2 generates slightly better pseudo-labels along the thin and small objects.

\begin{figure}[!t]
  \centering
  \begin{tabular}{ c c c c c}
    \includegraphics[width=0.19\textwidth]{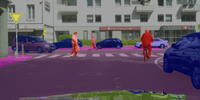} &
    \includegraphics[width=0.19\textwidth]{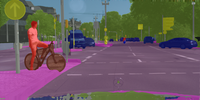} &
    \includegraphics[width=0.19\textwidth]{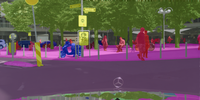} &
    \includegraphics[width=0.19\textwidth]{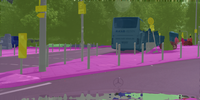} &
    \includegraphics[width=0.19\textwidth]{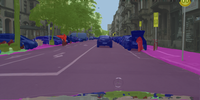} \\
    \includegraphics[width=0.19\textwidth]{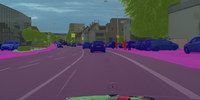} &
    \includegraphics[width=0.19\textwidth]{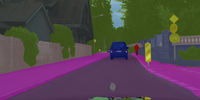} &
    \includegraphics[width=0.19\textwidth]{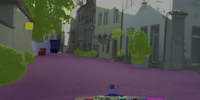} &
    \includegraphics[width=0.19\textwidth]{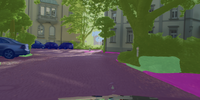} &
    \includegraphics[width=0.19\textwidth]{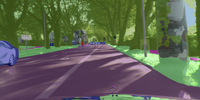} \\
    \includegraphics[width=0.19\textwidth]{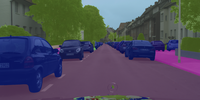} &
    \includegraphics[width=0.19\textwidth]{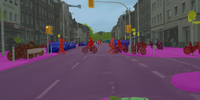} &
    \includegraphics[width=0.19\textwidth]{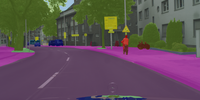} &
    \includegraphics[width=0.19\textwidth]{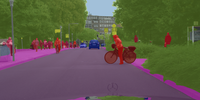} &
    \includegraphics[width=0.19\textwidth]{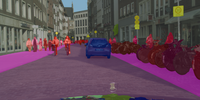}
  \end{tabular}
  \caption{{\bf Segmentation results by Student on Cityscapes val set.}
  }
  \label{fig:cityscapes_sample_overlays}
\end{figure}

{\bf Visualization of segmentation results:} In \figref{fig:cityscapes_sample_overlays}, we visualize some segmentation results obtained by the Student network on val-fine set.

\subsection{Ablation Studies}

In this subsection, we provide ablation studies on several design choices. Xception-71 is used as the backbone if not specified.

{\bf Training iterations:} First, we verify that the performance improvement does not solely result from longer training iterations, but from the extra large pseudo-labeled images. We train Panoptic-DeepLab \cite{cheng2019panoptic} with  60K iterations on Cityscapes train-fine set and obtain a PQ of 62.9\%. We increase the training iterations to 120K iterations, but do not observe any improvement (62.7\% PQ) (\ie, performance saturates after 60K iterations). On the other hand, our proposed Naive-Student attains a better performance with 180K iterations (65.3\% PQ) when trained with the larger train-sequence set.

{\bf Design choices for the Teacher to generate pseudo-labels:} When generating the pseudo-labels, there are four factors involved in our design, namely (1) assignment of void label to the ego-car region, (2) employment of test-time augmentation, (3) more Cityscapes human-labeled images, and (4) Mapillary Vistas pretraining. This ``Void-Ego-Car'' design, or factor (1), improves 0.7\% PQ, 0.5\% AP, and 0.4\% mIOU. We think the wrongly generated labels in the ego car region slightly affect the model training. Without employing the test-time augmentation, (\ie, multi-scale inference and left-right flipping), when generating the pseudo-labels, the performance drops by 0.8\% PQ, 1.1\% AP, and 0.9\% mIOU. Excluding more Cityscapes human-labeled images for fine-tuning the Teacher network degrades the performance by 1.4\% PQ, 1.2\% AP, and 1.3\% mIOU. Finally, if the Teacher network is not pretrained on the Mapillary Vistas dataset, the performance decreases by 2.2\% PQ, 2.2\% AP, and 1.5\% mIOU.

\begin{table*}[!t]
  \centering
  \caption{ {\bf Design choices for the Teacher to generate pseudo-labels.}  {\bf Void-Ego-Car:} Assignment of void label to ego-car regions. {\bf Test-Aug:} Test-time augmentation (\ie, multi-scale inputs and left-right flips). {\bf Val-Fine:} Inclusion of Cityscapes val-fine set for training the Teacher network. {\bf MV-Pretrained:} Employment of a pretrained checkpoint on Mapillary Vista for the Teacher network. Results presented on Cityscapes validation set. Note the Student network has no access to the validation set.}
  \scalebox{0.8}{
  \begin{tabular}{c c c c | c c c }
    \toprule[0.2em]
    \multicolumn{4}{c}{Pseudo-Label Generation Scheme} &  \multicolumn{3}{c}{Validation Set Results}\\
    Void-Ego-Car & Test-Aug & Val-Fine & MV-Pretrained & PQ (\%) & AP (\%) & mIOU (\%) \\
    \toprule[0.2em]
    \cmark & \cmark & \cmark & \cmark & 67.5 & 39.8 & 83.7  \\
           & \cmark & \cmark & \cmark & 66.8 & 39.3 & 83.3  \\
    \cmark &        & \cmark & \cmark & 66.7 & 38.7 & 82.8  \\
    \cmark & \cmark &        & \cmark & 66.1 & 38.6 & 82.4  \\
    \cmark & \cmark & \cmark &        & 65.3 & 37.6 & 82.2  \\
    \bottomrule[0.1em]
  \end{tabular}
  }
  \label{tab:cityscapes_val_teacher}
\end{table*}

{\bf Design choices for training the Student:} In \tabref{tab:cityscapes_val_student}, we report the results when training the Student network with different training set splits. The baseline Student network, trained with Cityscapes train-fine, attains the performance of 63.1\% PQ, 35.2\% AP, and 80.1\% mIOU. Using the pseudo-labeled train-sequence, the performance is improved by 2.2\% PQ, 2.4\% AP, and 2.1\% mIOU. Mixing human-labeled train-fine and pseudo-labeled train-sequence slightly degrades the performance. We think it is because of the inconsistent annotations between human-labeled and pseudo-labeled images, since train-fine is a subset of train-sequence. Finally, adding more pseudo-labeled images (train-sequence and train-extra) improves the result to 66.9\% PQ, 40.2\% AP, and 84.2\% mIOU.

\begin{table*}[!t]
  \centering
  \caption{{\bf Design choices for training the Student.} We experiment with different training splits for training the Student. Results presented on Cityscapes validation set.
  }
  \scalebox{0.9}{
  \begin{tabular}{c c c | c c c }
    \toprule[0.2em]
    \multicolumn{3}{c}{Training Set for Student} &  \multicolumn{3}{c}{Validation Set Results}\\
    Train-Fine & Train-Sequence & Train-Extra &  PQ (\%) & AP (\%) & mIOU (\%) \\
    \toprule[0.2em]
    \cmark &        &        & 63.1 & 35.2 & 80.1  \\
           & \cmark &        & 65.3 & 37.6 & 82.2  \\
    \cmark & \cmark &        & 65.2 & 37.3 & 82.0  \\
           & \cmark & \cmark & 66.9 & 40.2 & 84.2  \\
    \bottomrule[0.1em]
  \end{tabular}
  }
  \label{tab:cityscapes_val_student}
\end{table*}

\begin{figure*}[!t]
    \centering
    \begin{tabular}{c c c}
        \includegraphics[width=0.32\textwidth]{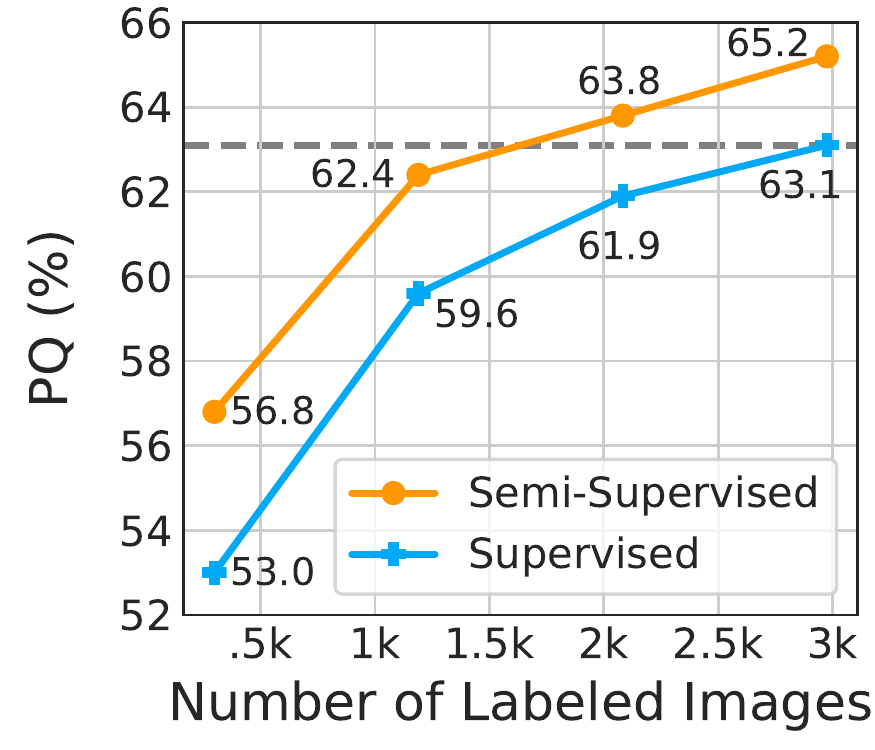} &
        \includegraphics[width=0.32\textwidth]{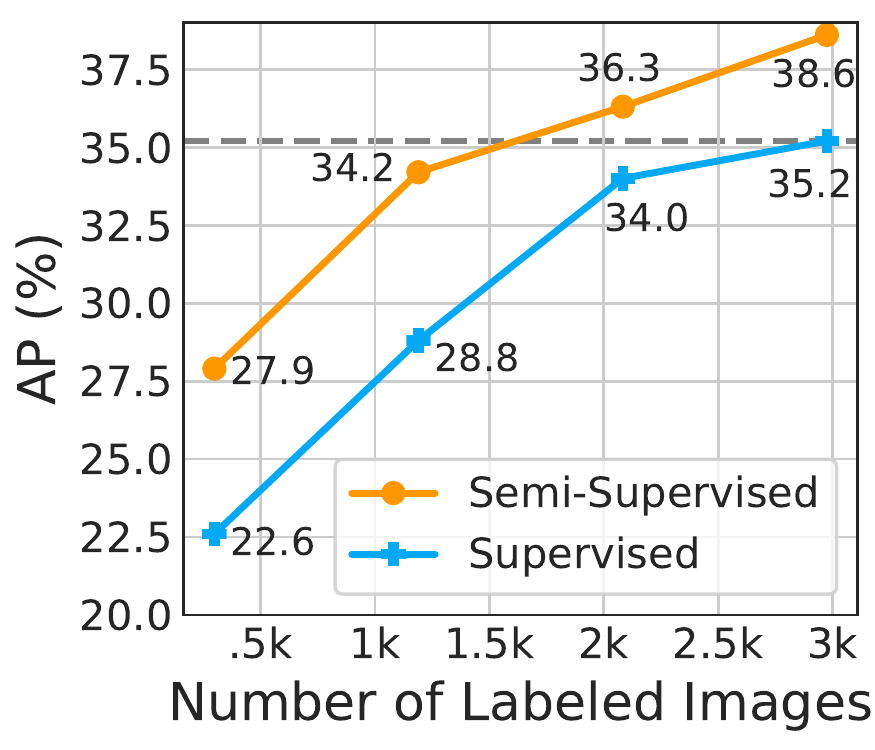} &
        \includegraphics[width=0.32\textwidth]{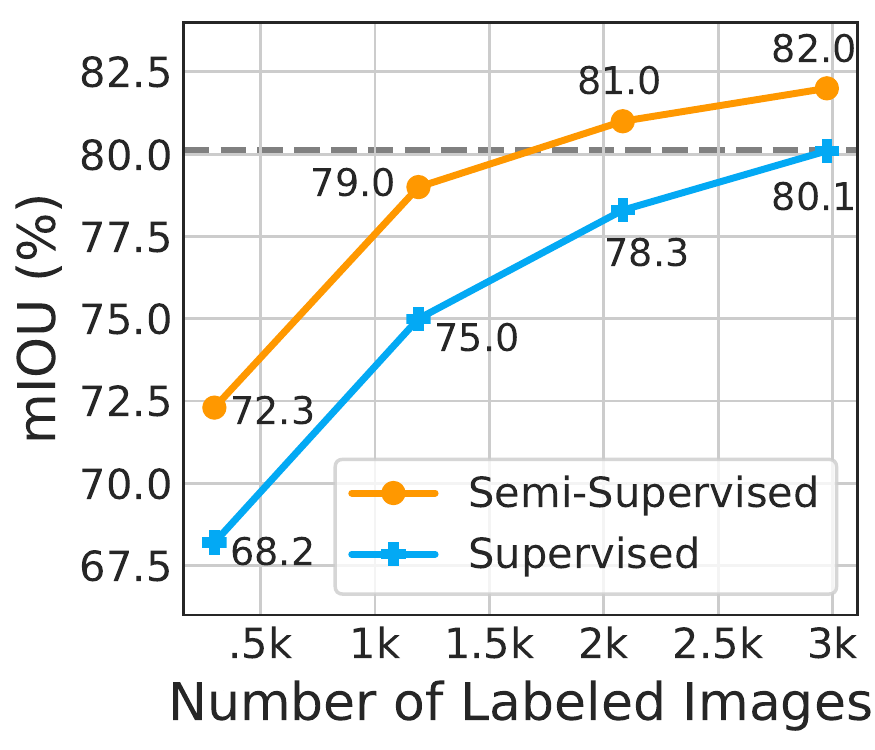} \\
        (a) PQ & (b) AP & (c) mIOU \\
    \end{tabular}
    \caption{{\bf Semi-supervised learning with a fraction of the original labels may match supervised segmentation performance.} Results presented on Cityscapes validation data. We vary the numbers of human-annotated images from train-fine set. With only 40\% of labeled data, our semi-supervised learning method attains 98.9\% PQ, 97.2\% AP, and 98.6\% mIOU of the performance of their fully supervised counterparts.}
    \label{fig:labeled_size}
\end{figure*}

{\bf Vary human-labeled images and fix pseudo-labeled images:} In \figref{fig:labeled_size}, we explore the semi-supervised setting with different amounts of human-labeled images but fixed amount of pseudo-labeled images. In particular, the Teacher network has {\it only} been trained with different numbers of Cityscapes train-fine images (\ie, no other human-labeled images, such as Mapillary Vistas). The generated pseudo-labels (on Cityscapes train-sequence and train-extra) are used to train another Student network. Both Teacher and Student networks employ the Xception-71 as backbone. For comparison, we also show the performance of the supervised setting with the same amount of human-labeled images. As shown in the figure, we observe (1) the semi-supervised learning setting consistently improves over the fully supervised setting in all three metrics (PQ, AP, and mIOU) as more human-labeled images are exploited, (2) when using only 40\% of the human-labeled images, our semi-supervised learning method could reap 98.9\%, 97.2\%, and 98.6\% performance from its fully supervised counterparts in PQ, AP, and mIOU, respectively, and (3) when using 100\% of the human-labeled images, our semi-supervised learning method attains 65.2\% PQ, 38.6\% AP, and 82\% mIOU, comparable to the fully supervised counterpart with a Mapillary Vistas pretrained checkpoint (65.3\% PQ, 38.8\% AP, and 82.5\% mIOU in \cite{cheng2019panoptic}).

\begin{figure*}[!t]
    \centering
    \begin{tabular}{c c c}
        \includegraphics[width=0.32\textwidth]{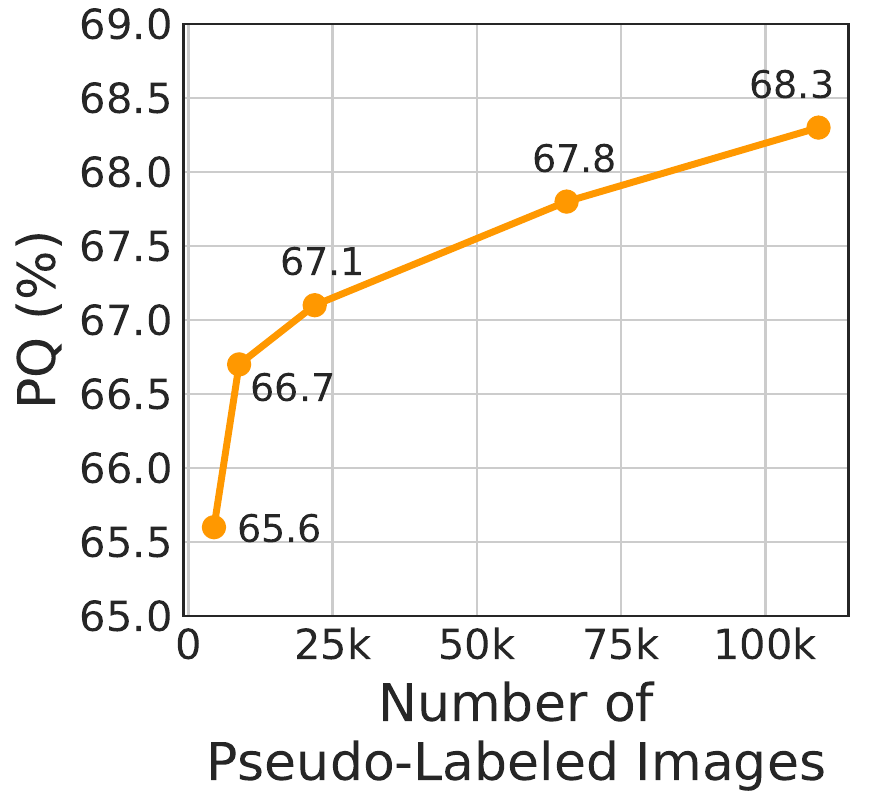} &
        \includegraphics[width=0.32\textwidth]{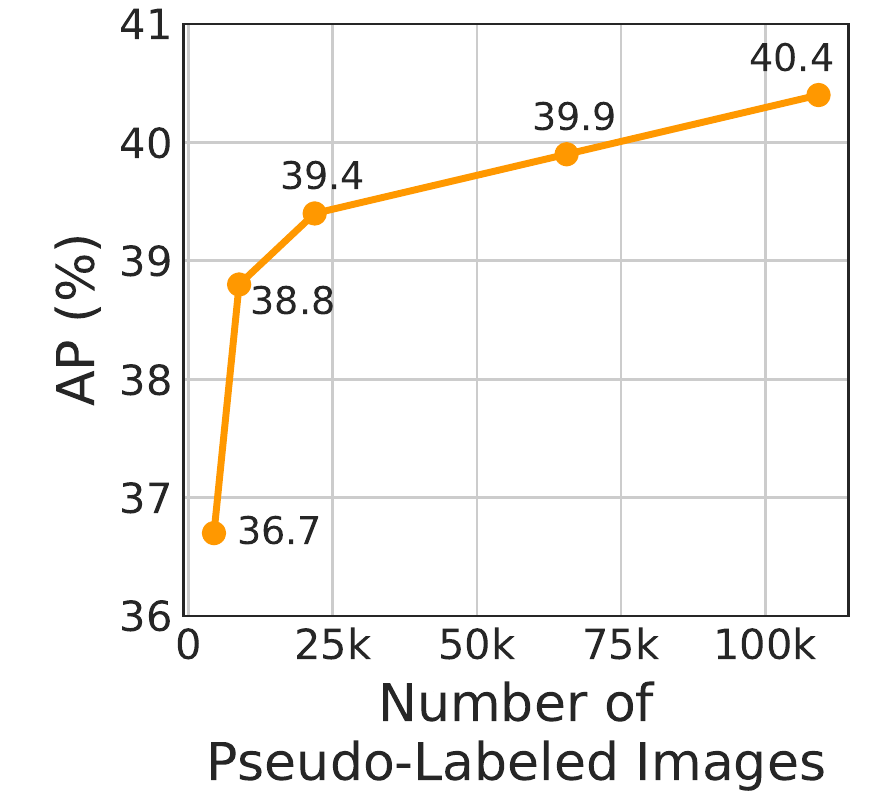} &
        \includegraphics[width=0.32\textwidth]{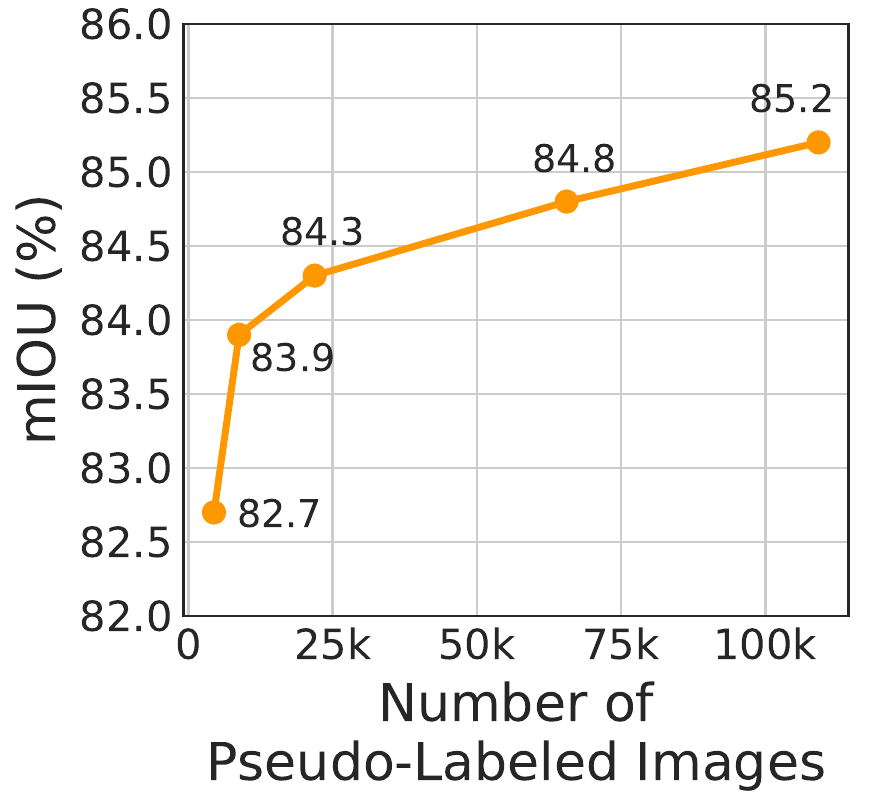} \\
        (a) PQ & (b) AP & (c) mIOU \\
    \end{tabular}
    \caption{{\bf Increasing the amount of unlabeled data improves segmentation performance} for PQ, AP, and mIOU. Results presented on Cityscapes validation data.}
    \label{fig:video_size}
\end{figure*}

{\bf Fix human-labeled images and vary pseudo-labeled images:} In \figref{fig:video_size}, we explore the semi-supervised setting with different amounts of pseudo-labeled images. In particular, the Teacher network will generate different numbers of pseudo-labeled images for training the Student network. As shown in the figure, we observe consistent improvement in all three metrics when more and more pseudo-labeled images are included in the training.

{\bf Training method:} In \tabref{tab:cityscapes_iterative_ss}, we experiment with the effect of different training methods: supervised, semi-supervised, and iterative semi-supervised learning. We employ our most powerful backbone, WR-41,  attempting to push the envelope of performance. We observe a significant improvement of semi-supervised learning over supervised learning by 5.1\% PQ, 4.4\% AP, and 5.2\% mIOU, mostly because of the small Cityscapes dataset. Adopting the iterative semi-supervised learning further improves the performance by 1.3\% PQ, 1\% AP, and 0.2\% mIOU. We think there is more room for improving  PQ and AP, since the mIOU result is starting to be saturated, as demonstrated in the public leader-board (\ie, differences between top-performing models are about 0.1\%).

\begin{table*}[!t]
  \centering
  \caption{{\bf Comparisons with different training methods.} Results presented on Cityscapes validation set. The proposed WR-41 is used as the network backbone. Semi-supervised learning (\ie, only iterate once in our framework) significantly improves the performance, while iterative semi-supervised learning further improves the results.
  }
  \scalebox{0.8}{ %
  \begin{tabular}{c | c c c }
    \toprule[0.2em]
    Training Method & PQ (\%) & AP (\%) & mIOU (\%) \\
    \toprule[0.2em]
    supervised & 64.0 & 38.0 & 80.7 \\
    semi-supervised & 69.1 & 42.4 & 85.9 \\
    {\it iterative} semi-supervised & 70.4 & 43.4 & 86.1 \\
    \bottomrule[0.1em]
  \end{tabular}
  } \\
  \label{tab:cityscapes_iterative_ss}
\end{table*}

{\bf Transfer learning from Cityscapes to Mapillary Vistas:} The transfer learning from the large-scale Mapillary Vistas to Cityscapes has been shown to be effective in the literature \cite{zhu2019improving,porzi2019seamless,cheng2019panoptic} and in our work as well, since both datasets contain street-view images. In \tabref{tab:student_transfer}, we experiment with the other direction of transfer learning from Cityscapes to Mapillary Vistas. The baseline model with WR-41 backbone, pretrained only on ImageNet \cite{ILSVRC15}, attains the performance of 37.1\% PQ, 16.7\% AP, and 56.2\% mIOU. If we pretrain the model on the original Cityscapes trainval-fine set, we observe a slight degradation. Interestingly, when we further pretrain the model on the generated pseudo-labels, we observe a small amount of 0.7\% improvement in PQ. We think the improvement gained from Cityscapes pretraining is marginal, mainly because the Cityscapes images are mostly taken in Germany, while Mapillary Vistas contains more diverse images.

\begin{table*}[!t]
  \centering
  \caption{ {\bf Transfer learning from Cityscapes to Mapillary Vistas.} Results presented on Mapillary Vistas validation data. We experiment with the network pretrained on  ImageNet \cite{ILSVRC15}, Cityscapes \cite{cordts2016cityscapes} labeled data (trainval-fine set), and Cityscapes pseudo-labeled data (trainval-sequence, and train-extra).}
  \scalebox{0.8}{
  \begin{tabular}{c c |  c  c c}
    \toprule[0.2em]
    \multicolumn{2}{c}{Training Set} & \multicolumn{3}{c}{Val Set} \\
    Labeled & Pseudo-Labeled & PQ (\%) & AP (\%) & mIOU (\%) \\
    \toprule[0.2em]
           &        & 37.1 & 16.7 & 56.2 \\
    \cmark &        & 36.9 & 16.5 & 55.3 \\           
    \cmark & \cmark & 37.8 & 17.0 & 56.2 \\
    \bottomrule[0.1em]
  \end{tabular}
  }
  \label{tab:student_transfer}
\end{table*}

\subsection{Modified Wide ResNet-38: WR-41}

In this subsection, we report the experimental results with our modified Wide ResNet-38 \cite{wrn2016wide,wu2019wider}, called WR-41, on both ImageNet \cite{ILSVRC15} and Cityscapes \cite{cordts2016cityscapes}.

{\bf ImageNet-1K val set:} In \tabref{tab:imagenet}, we report the results on the ImageNet-1K validation set. As shown in the table, our TensorFlow re-implementation of wide ResNet-38 (WR-38), proposed in \cite{wu2019wider}, attains 20.36\% Top-1 error, which is slightly worse than the one reported in the original paper. We think there are some differences between the deep learning libraries. Note however that our main focus is on the segmentation results, while ImageNet is only used for pretraining. Our proposed WR-41 achieves a slightly better performance. Employing the drop path \cite{huang2016deep} with a constant survival probability 0.8 improves the performance.

\begin{table}[!t]
  \centering
  \caption{{\it Single-model} error rates on ImageNet-1K validation set.}
  \scalebox{0.8}{
  \begin{tabular}{c | c | c  c }
    \toprule[0.2em]
    Backbone & Drop-Path & Top-1 Error & Top-5 Error \\
    \toprule[0.2em]
    WR-38 \cite{wu2019wider} (our TF imp.) &        & 20.36\% & 5.11\%  \\
    WR-38 \cite{wu2019wider} (our TF imp.) & \cmark & 19.89\% & 4.98\%  \\
    \midrule
    WR-41                                  &        & 20.08\% & 4.93\%  \\
    WR-41                                  & \cmark & 19.41\% & 4.68\%  \\
    \bottomrule[0.1em]
  \end{tabular}
  }
  \label{tab:imagenet}
\end{table}

\begin{table}[!t]
  \centering
  \caption{Adopting Panoptic-DeepLab with our WR-41 achieves better segmentation accuracy on Cityscapes val set with fewer model parameters and fewer M-Adds than with our TensorFlow re-implemented WR-38.}
  \scalebox{0.7}{
  \begin{tabular}{c | c c | c c | c c c}
    \toprule[0.2em]
    Backbone & Drop-Path & Multi-Grid & Params (M) & M-Adds (B) & PQ (\%) & AP (\%) & mIOU (\%) \\
    \toprule[0.2em]
    WR-38 \cite{wu2019wider} (our TF imp.) &        &        & 173.75 & 3486.32 & 62.4 & 35.3 & 79.1 \\
    WR-38 \cite{wu2019wider} (our TF imp.) & \cmark &        & 173.75 & 3486.32 & 62.6 & 36.7 & 79.5 \\
    WR-38 \cite{wu2019wider} (our TF imp.) & \cmark & \cmark & 173.75 & 3493.80 & 63.1 & 37.4 & 80.1 \\
    \midrule
    WR-41 &        &        & 147.27 & 3238.81 & 63.0 & 35.5 & 80.0 \\
    WR-41 & \cmark &        & 147.27 & 3238.81 & 63.6 & 36.8 & 80.4 \\
    WR-41 & \cmark & \cmark & 147.27 & 3246.40 & 64.0 & 38.0 & 80.7 \\
    \bottomrule[0.1em]
  \end{tabular}
  }
  \label{tab:cityscapes_val_wr38}
\end{table}

\begin{table}[!t]
  \centering
  \caption{{\bf Effect of test-time augmentation on Cityscapes val set.} {\bf MV}: Mapillary Vistas pretrained. {\bf Flip}: Left-right flips. {\bf MS:} Multi-scale inputs.
  }
  \scalebox{0.8}{
  \begin{tabular}{c | c c c | c c c}
    \toprule[0.2em]
    Method & MV & Flip & MS & PQ (\%) & AP (\%) & mIOU (\%) \\
    \toprule[0.2em]
    Panoptic-DeepLab (WR-41) &        &        &        & 64.0 & 38.0 & 80.7  \\
    Panoptic-DeepLab (WR-41) &        & \cmark &        & 64.5 & 39.3 & 81.2 \\
    Panoptic-DeepLab (WR-41) &        & \cmark & \cmark & 65.0 & 40.7 & 81.4 \\
    \midrule
    Panoptic-DeepLab (X-71) \cite{cheng2019panoptic} &        & \cmark & \cmark & 64.1 & 38.5 & 81.5 \\
    \midrule \midrule
    Panoptic-DeepLab (WR-41) & \cmark &        &        & 66.5 & 41.5 & 83.4  \\
    Panoptic-DeepLab (WR-41) & \cmark & \cmark &        & 66.7 & 41.8 & 83.7 \\
    Panoptic-DeepLab (WR-41) & \cmark & \cmark & \cmark & 67.3 & 43.4 & 83.8 \\
    \midrule
    Panoptic-DeepLab (X-71) \cite{cheng2019panoptic} & \cmark & \cmark & \cmark & 67.0 & 42.5 & 83.1 \\
    \bottomrule[0.1em]
  \end{tabular}
  }
  \label{tab:cityscapes_val_wr38_aug}
\end{table}

\begin{table}[!t]
  \centering
  \caption{{\bf Panoptic-DeepLab with proposed WR-41 backbone on Cityscapes test set.} {\bf MV:} Mapillary Vistas pretrained.
  }
  \scalebox{0.8}{
  \begin{tabular}{c | c | c c c}
    \toprule[0.2em]
    Method & MV & PQ (\%) & AP (\%) & mIOU (\%) \\
    \toprule[0.2em]
    Panoptic-DeepLab (X-71) \cite{cheng2019panoptic} &        & 62.3 & 34.6 & 79.4 \\
    
    Panoptic-DeepLab (WR-41) (ours) & & 63.7 & 36.5 & 81.5 \\
    \midrule
    Panoptic-DeepLab (X-71) \cite{cheng2019panoptic} & \cmark & 65.5 & 39.0 & 84.2 \\
    
    Panoptic-DeepLab (WR-41) (ours) & \cmark & 66.5 & 40.6 & 84.5 \\
    
    \bottomrule[0.1em]
  \end{tabular}
  }
  \label{tab:cityscapes_test_wr38_aug}
\end{table}

{\bf Cityscapes val set:} In \tabref{tab:cityscapes_val_wr38}, we report the Cityscapes validation set results when using Panoptic-DeepLab \cite{cheng2019panoptic} with WR-38 (our TensorFlow re-implementation) and WR-41 as backbones. As shown in the table, we observe (1) using drop path \cite{huang2016deep} (constant survival probability 0.8) consistently improves the performance in both backbones, (2) the performance could be further improved by adopting the multi-grid scheme proposed in \cite{chen2017deeplabv3} (where the unit rates in the last two or three residual blocks are set to (1, 2) or (1, 2, 4) for WR-38 and WR-41, respectively), (3) using WR-41 as backbone slightly improves over WR-38, and (4) Panoptic-DeepLab with WR-41 as backbone is slightly faster (and with slightly fewer parameters) than with WR-38 because the ASPP module \cite{chen2018deeplabv2} is added on the last feature map with 2048 channels (instead of 4096 channels). Additionally, the GPU inference times (Tesla V100-SXM2) on a $1025\times2049$ input for WR-38 and WR-41 are 437.9 ms and 396.5 ms, respectively.

In \tabref{tab:cityscapes_val_wr38_aug}, we report the effect of using test-time augmentation (\ie, multi-scale inputs and left-right flips) and pretraining on Mapillary Vistas, when using Panoptic-DeepLab with WR-41 as network backbone. The performance consistently improved with test-time augmentation and pretraining on Mapillary Vistas. Additionally, adopting Panoptic-DeepLab with WR-41 slightly outperforms Panoptic-DeepLab with X-71 as reported in \cite{cheng2019panoptic}.

{\bf Cityscapes test set:} In \tabref{tab:cityscapes_test_wr38_aug}, we report the Cityscapes test set results when using Panoptic-DeepLab \cite{cheng2019panoptic} with our modified WR-41. Without extra data, our Panoptic-DeepLab (WR-41) outperforms Panoptic-DeepLab (X-71) by 1.4\% PQ, 1.9\% AP, and 2.1\% mIOU. With Mapillary Vistas pretraining, our Panoptic-DeepLab (WR-41) outperforms Panoptic-DeepLab (X-71) by 1.0\% PQ and 1.6\% AP, and 0.3\% mIOU.

%% file: sections/5.conclusion.tex
\section{Conclusion}
\label{sec:conclusion}
In this work, we have described an iterative semi-supervised learning method that significantly improves the performance of urban scene segmentation on Cityscapes, simultaneously tackling semantic, instance, and panoptic segmentation. This semi-supervised learning procedure effectively harnesses both unlabeled video frames and extra unlabeled images to improve the predictive performance of the model without the creation of additional architectures and learned modules. Namely, pseudo-labeled data garnered through a simple data augmentation (\ie, multi-scale inputs and left-right flips) suffices to boost performance on supervised learning tasks. As a result, our model sets the new state-of-art performance at all three Cityscapes benchmarks without the need to fine-tune or any special design on each task. We hope our simple yet effective learning scheme could establish a baseline procedure to harness the abundant unlabeled video sequences and extra images for computer vision tasks.

\paragraph{Acknowledgments}
We would like to thank the support from Google Mobile Vision and Brain.